\documentclass[runningheads]{llncs}

\usepackage{eccv}

\usepackage{eccvabbrv}

\usepackage{graphicx}
\usepackage{booktabs}

\usepackage[accsupp]{axessibility}  

\usepackage{hyperref}

\usepackage{orcidlink}

\begin{document}

\title{UnrealEvent: Synthetic Dataset Generation and Annotation for Event-Based Vision} 

\titlerunning{UnrealEvent}

\author{Yentel Schoovaerts\and
Zijie Ning\orcidlink{0009-0006-6117-3302} \and
Patrick Vandewalle\orcidlink{0000-0002-7106-8024}}

\authorrunning{Y. Schoovaerts et al.}

\institute{EAVISE-PSI, ESAT, KU Leuven, Belgium\\
\email{zijie.ning@kuleuven.be}}

\maketitle

\begin{abstract}
  Event-based cameras provide high temporal resolution, low power consumption, and a high dynamic range, but progress in event-based object detection remains constrained by the scarcity of annotated training data. To solve this issue, this paper introduces the tool UnrealEvent, a synthetic dataset-generation pipeline implemented in Unreal Engine 5.5. The pipeline combines a V2E-inspired event-frame generator with EasySynth semantic-rendering passes, allowing rendered level sequences to be converted into paired event representations and YOLO-format bounding-box annotations without manual labeling. Events are generated from per-pixel log-luminance differences between consecutive frames, with contrast thresholds sampled independently from a normal distribution to emulate the fabrication-induced threshold mismatch of real sensors. Using this pipeline, we provide a synthetic dataset and evaluate YOLOv8 detectors under synthetic, zero-shot real, and real fine-tuning training on eTraM. The results show strong in-domain synthetic performance but a substantial synthetic-to-real gap, especially for pedestrians. Fine-tuning on a small real subset improves real-domain performance, while the remaining gap highlights the importance of asset diversity and sensor calibration.
\end{abstract}

\section{Introduction}
\label{sec:intro}

Object detection is a fundamental task in computer vision, aiming to identify objects of interest in images and estimate their spatial properties such as position, scale, and orientation. Driven by advances in sensing, algorithms, and computing, it has been adopted across automation, surveillance, wildlife conservation, and healthcare \cite{pagire_comprehensive_2025}. Most of this progress, however, relies on conventional frame-based cameras, which sample the entire scene at fixed temporal intervals. Despite their effectiveness, frame-based sensors suffer from high latency, limited dynamic range, high power consumption, and motion blur when capturing fast-moving objects, limitations that are especially problematic in dynamic or resource-constrained settings \cite{cazzato_application-driven_2024}.\\
Event-based cameras, also known as neuromorphic vision sensors, have emerged as an alternative inspired by the biological visual system. Rather than acquiring full frames synchronously, each pixel operates independently and asynchronously, generating an event only when its local change in luminance exceeds a threshold. This paradigm yields very high temporal resolution without motion blur, a sparse and data-efficient stream with correspondingly low bandwidth and power requirements, and a high dynamic range achieved by responding to logarithmic intensity changes. Consequently, event cameras remain robust in challenging conditions such as nighttime scenes or abrupt illumination changes where conventional sensors often fail\cite{chakravarthi_recent_2024}.\\
Despite these advantages, the adoption of event-based vision remains limited, largely due to the scarcity of large-scale, well-annotated datasets. Neuromorphic sensors are costly, and remain confined to specialized research communities \cite{annamalai2020evanneuromorphiceventbasedanomaly}, so publicly available event-based datasets are at least an order of magnitude scarcer than their frame-based counterparts. The asynchronous, discrete nature of event data further complicates its integration into standard learning pipelines, typically requiring conversion into frame-like representations that may discard fine-grained temporal information \cite{rebecq2019highspeedhighdynamic, Qu_2024}. Event-based dataset creation also inherits the general difficulty of capturing target objects under diverse illumination, environmental, and safety-critical conditions, making the process logistically complex and expensive.\\
These factors have motivated a shift toward synthetic data generation as a scalable and controlled alternative\cite{kapania2025examiningexpandingrolesynthetic}. Synthetic data preserves the underlying structure of real-world data while enabling large-scale generation without specialized hardware, eliminating the safety concerns of recording hazardous scenarios\cite{chang2024safesimsafetycriticalclosedlooptraffic} and avoiding privacy issues since no real individuals or locations are captured\cite{guo2024generativeaisyntheticdata}. Such data is typically produced with rendering engines such as Blender or Unreal Engine (UE), or dedicated simulators such as ESIM\cite{rebecq_esim_2018}, DAVIS\cite{mueggler_event-camera_2017}, and V2E\cite{hu_v2e_2021}. UE is particularly attractive for its near-photorealistic output and its support for interactive navigation and camera trajectory control, enabling flexible and repeatable data generation, albeit at higher computational cost\cite{rebecq_esim_2018}.\\
Motivated by the dataset scarcity that constrains event-based object detection, this work investigates synthetic data generation within simulation environments. Specifically, we address three research questions: (1) how an Unreal Engine pipeline can generate event-like visual representations and synchronized annotations; (2) how a synthetically trained event-based object detection model generalizes to real-world data; and (3) how an automated pipeline can be built to generate annotated data. In doing so, we aim to contribute to more accessible and scalable methods for event-based dataset creation.\\

\section{Related Work}
\label{sec:related}
\subsection{Event Data Simulation}
\begin{table}[h!]
\centering
\caption{Overview of simulators}
\label{simulators-table}
\renewcommand{\arraystretch}{1.3}
\begin{tabular}{p{2cm} p{2.5cm} p{3cm} p{3cm} p{0.8cm}}
\toprule
\multicolumn{5}{c}{\textbf{Event-based data simulators}} \\
\midrule
\textbf{Simulator} & \textbf{Noise} & \textbf{Method} & \textbf{Defining feature} & \textbf{Year} \\
\midrule
DAVIS Simulator~\cite{mueggler_event-camera_2017} & 
Ideal Pixel &
Based on Blender &
Linear Time Interpolation Sampling &
2017 \\

ESIM~\cite{rebecq_esim_2018} & 
Standard Noise Model &
Based on Rendering Engine &
Adaptive Sampling &
2018 \\

V2E~\cite{hu_v2e_2021} & 
Temporal Noise, Leak Events, Finite Bandwidth, Gaussian Threshold Distribution  &
RGB-video to Event Sequence &
Inclusion of Motion Blur Effects &
2021 \\

DVS-Voltmeter~\cite{lin_dvs-voltmeter_2022} & 
Stochastic Process &
Stochastic Process-based Event Simulator  &
Incorporating Sensor Physics &
2022 \\

V2CE~\cite{zhang_v2ce_2024} & 
Learned by Training on Real Event Voxels &
RGB-video to Event Sequence &
Local Dynamics-Aware Timestamp Inference &
2024 \\
\bottomrule
\end{tabular}
\end{table}
Because real event data is costly to acquire, simulation has become a central tool for generating event streams under controlled, repeatable conditions. The DAVIS simulator\cite{mueggler_event-camera_2017} was among the first, rendering a virtual scene along a predefined camera trajectory with Blender and reconstructing event timings through linear temporal interpolation, under the constraint that inter-frame motion stays below one third of a pixel. While effective as an initial step, it models only an idealized pixel and omits the sensor noise present in physical cameras. ESIM\cite{rebecq_esim_2018} advanced this by tightly coupling the simulator to the rendering engine and adaptively adjusting the sampling rate to the dynamics of the visual signal, yielding more faithful event timing. It also introduced a standard noise model motivated by the observation that the contrast threshold varies approximately according to a normal distribution. Residual discrepancies nonetheless remain, as simple noise models do not capture effects such as missed events during the pixel refractory period at high event rates\cite{brandli_real-time_2014}.\\
Subsequent simulators target more realistic, non-ideal behavior. V2E\cite{hu_v2e_2021} operates directly on conventional RGB video rather than a 3D scene, modeling pixel-level Gaussian threshold mismatch, finite intensity-dependent bandwidth, intensity-dependent noise, and motion-blur effects, making it well suited to converting existing frame-based recordings into events. DVS-Voltmeter\cite{lin_dvs-voltmeter_2022} grounds event generation in sensor physics, jointly modeling circuit voltage variations, the stochastic nature of photon reception, and temperature- and parasitic-current-induced noise\cite{nozaki_temperature_2017} as stochastic processes. More recently, V2CE\cite{zhang_v2ce_2024} improves timestamp recovery from event voxels via local dynamics-aware inference, reporting substantially lower error and event counts much closer to real sensors than prior video-to-event methods. Table 1 summarizes these simulators.\\
Despite steady gains in realism, all of these simulators share one limitation relevant to dataset creation: they produce event streams but no annotations. This restricts their direct use for training supervised models such as object detectors, and motivates coupling event simulation with automatic annotation, which is the focus of this work.\\

\subsection{Event-Based Datasets}
High-quality datasets are a prerequisite for training and benchmarking event-based vision models, yet they remain scarce relative to datasets for frame-based vision. Real-world datasets are captured with physical event cameras: MVSEC\cite{8288670} provides stereo sequences with depth and pose ground truth from fused IMU, LiDAR, motion-capture, and GPS; DSEC\cite{Gehrig21ral} targets autonomous driving with LiDAR-derived depth refined through mapping; and EventVOT\cite{wang_event_2023} and eTraM\cite{verma_etram_2024} use high-definition Prophesee sensors for tracking and traffic monitoring, with professionally annotated bounding boxes. Such datasets achieve high realism but are expensive and logistically demanding to produce, requiring extensive recording across conditions and costly manual or professional annotation.\\
A recurring theme across these datasets is that annotation is method-specific and tightly bound to the acquisition setup: MVSEC and DSEC rely on LiDAR and GPS for depth ground truth, SeACT\cite{zhou_exact_2024} uses GPT-4 to generate textual action descriptions, the Yelan synthetic dataset\cite{ZHANG2023126388} derives joint keypoints from its underlying Blender models, and SEVD\cite{aliminati2024sevd} uses CARLA's bounding-box API. SEVD is the closest prior work to ours, being the first synthetic event dataset to use multiple event sensors in CARLA\cite{dosovitskiy_carla_2017} with automatically generated 2D and 3D bounding boxes. However, CARLA is built specifically for autonomous-driving research, so SEVD inherits that scope and offers limited flexibility for other object categories or environments. This constraint motivates a general-purpose, annotation-aware pipeline that is not tied to a single application domain.\\

\subsection{Unreal Engine for Computer Vision}
Unreal Engine (UE) has become a widely used platform for synthetic data generation, owing to its high-performance, near-photorealistic rendering pipeline, which helps narrow the gap between real and synthetic data for training and evaluation\cite{articleYuJungKim}. Several tools build on UE to support computer vision (CV) workflows. UnrealCV was the first open-source plugin to connect UE to CV pipelines, exposing ground truth such as bounding boxes, depth maps, and segmentation masks, and enabling systematic generation of annotated image datasets\cite{qiu2016unrealcvconnectingcomputervision}. UnrealGT extends this to object-detection datasets, pairing image outputs (RGB, depth, semantic and instance segmentation, normals) with camera metadata to produce annotations automatically. ReplicANT\cite{Plum2023a} places 3D models into procedurally generated, configurable UE 5 environments and exports annotations for detection, tracking, pose estimation, and segmentation, while the Falling Things dataset\cite{tremblay2018fallingthingssyntheticdataset} uses a custom UE plugin to generate photorealistic images with pixel-accurate 2D/3D bounding boxes and segmentation masks.\\
These tools demonstrate that UE supports automatic, high-quality annotation for conventional frame-based vision. At the same time, ESIM\cite{rebecq_esim_2018} shows that UE can serve as a rendering backend for event simulation, but it provides no annotation capability. To our knowledge, no existing tool generates fully annotated event-based datasets within UE. This gap directly motivates our work: a simulation pipeline that converts rendered RGB video sequences into event representations while simultaneously producing synchronized bounding-box annotations, transferred from the semantic frames to the event data.\\

\section{Methodology}
\subsection{Simulation Environment}
The pipeline is built in UE 5.5, chosen for its photorealistic rendering and its compatibility with the open-source EasySynth plugin\cite{easysynth}, which renders annotation passes (semantic segmentation, depth, surface normals) from level sequences and supports custom post-processing effects. This extensibility lets us integrate the event-camera effect developed in this work directly into EasySynth's rendering, so that event frames and ground-truth labels are produced from the same scene. Vehicle and pedestrian assets were sourced from the Fab marketplace\cite{bugrimov_adventure_character, switchboardstudios_vehicle_variety_pack, leartesstudios_stylized_house_interior} for visual diversity. Level sequences are rendered at 300 fps; this high rate is essential, since faithful event simulation requires dense temporal sampling between frames, and it lets UE render directly at the target rate without the Super-SloMo interpolation that V2E applies to low-frame-rate video. The rate of 300 fps balances temporal fidelity against rendering time, as higher rates increase the frame count and thus generation cost.

\subsection{Event Generation}
\begin{figure}[h!]
\centering
\includegraphics[scale=0.5]{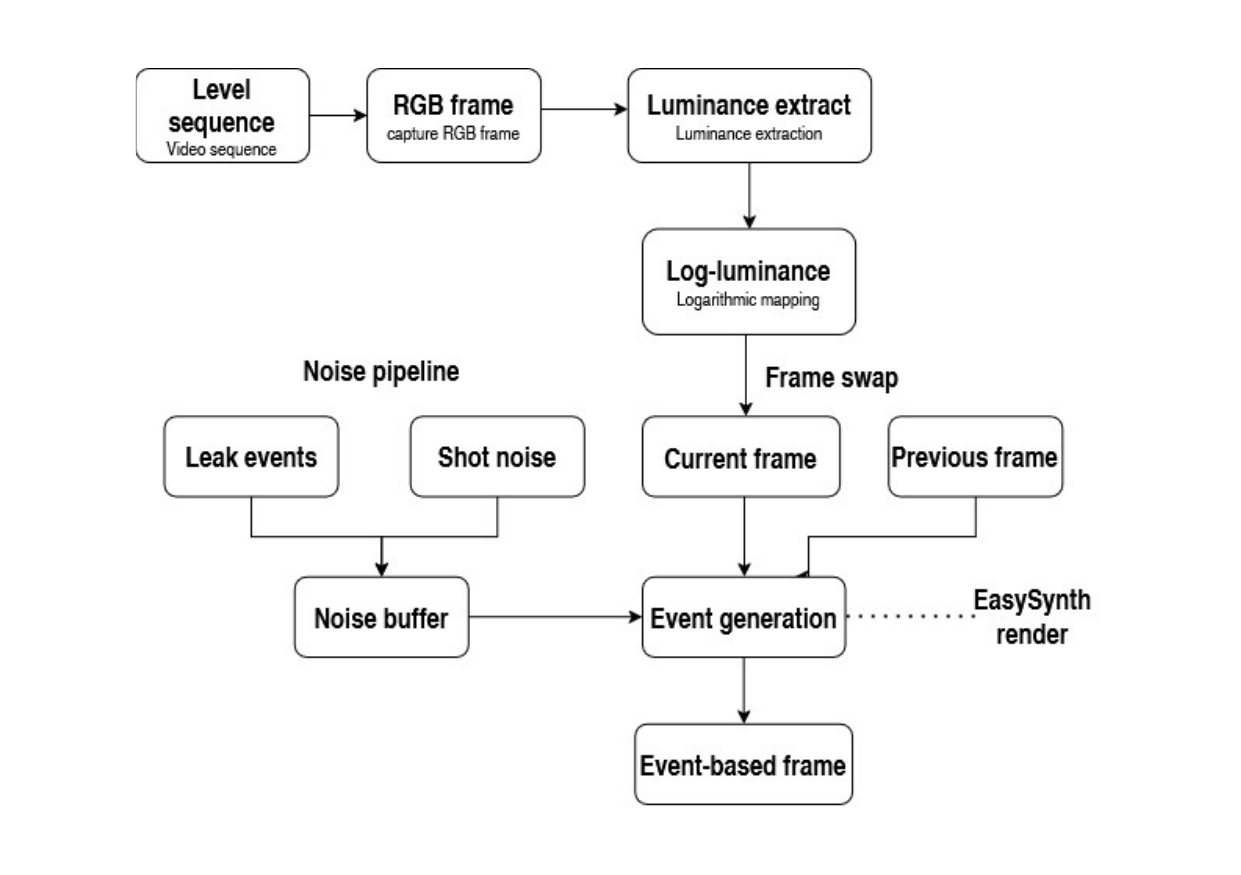}
\caption{Full simulation pipeline}
\label{fig:EventCamSimulationPipeline}
\end{figure}
\begin{figure}[h!]
    \centering
    \begin{subfigure}[b]{0.42\textwidth}
        \includegraphics[scale=0.30]{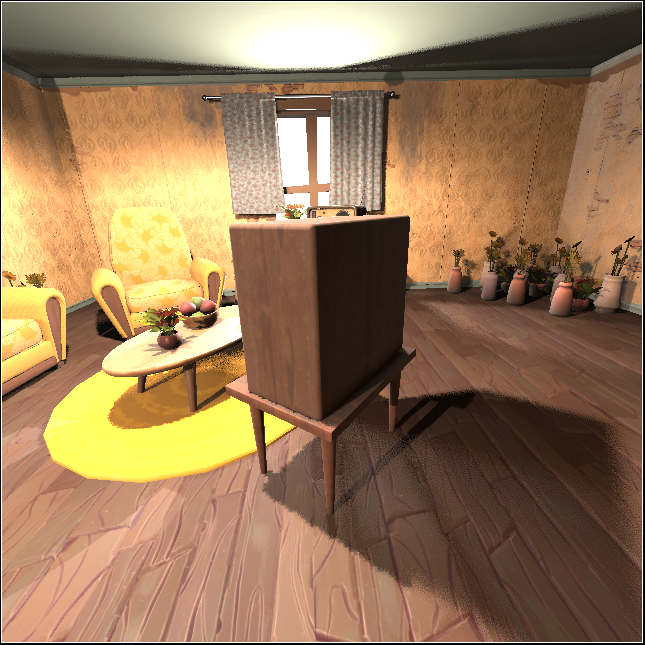}
        \label{fig:scene1_0a}
    \end{subfigure}
    \begin{subfigure}[b]{0.42\textwidth}
        \includegraphics[scale=0.25]{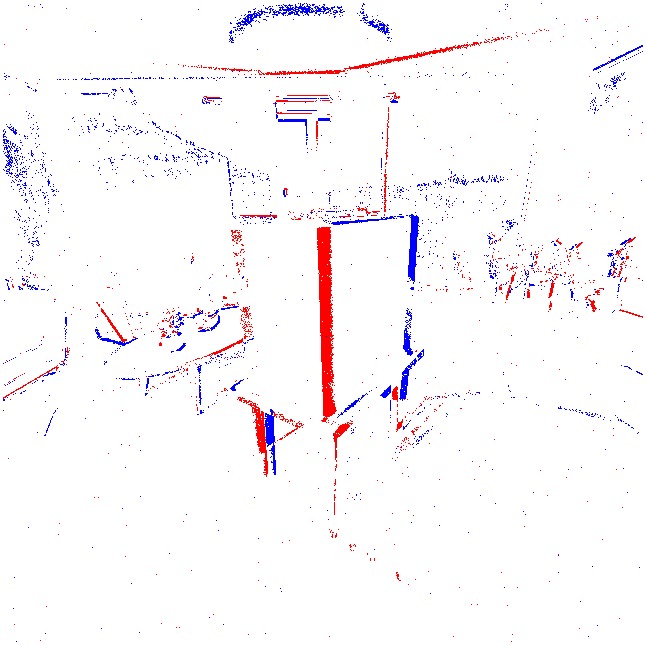}
        \label{fig:scene1_0b}
    \end{subfigure}
\caption{Example pipeline output: a rendered RGB frame (left) and its corresponding event representation (right). Blue and red denote ON ($\Delta L > 0$) and OFF ($\Delta L < 0$) events.}
\label{fig:scene1_a}
\end{figure}

The event-generation model is a V2E-inspired event-frame approximation adapted to operate on rendered frames inside UE. We deliberately adopt a rule-based rather than learning-based formulation: learning-based simulators such as EventGAN require paired event–image data that is itself scarce, and training on synthetic-only data yields a synthetic-to-synthetic domain gap with no guaranteed real-world transfer; moreover, the microsecond resolution of physical sensors is incompatible with the fixed render rates of game engines. A rule-based pipeline is therefore the more appropriate foundation, and the full pipeline is shown in Fig~\ref{fig:EventCamSimulationPipeline}.\\
Each rendered frame is first converted to luminance,
\begin{equation}
Y = 0.2126 \cdot R + 0.7152 \cdot G + 0.0722 \cdot B.
\end{equation}
and then mapped to the log-luminance domain $L=\log Y$, reflecting the logarithmic response that gives event cameras their high dynamic range. The core stage, implemented as a post-processing material operating on two consecutive log-luminance frames, computes the per-pixel temporal difference $\Delta L = L_t - L_{t-1}$. An ON (positive) event is assigned where $\Delta L>0$ and an OFF (negative) event where $\Delta L<0$. An event is emitted only if its magnitude exceeds a per-pixel contrast threshold; otherwise no event is written at that location for the current frame. Fig.~\ref{fig:scene1_a} shows an example of this output alongside its source RGB scene.\\
Rather than a single global threshold, each pixel draws its threshold independently from:
\begin{equation}
C_{pixel} = \mathcal{N}(C_{nominal}; 0.03).
\end{equation}
modeling the inter-pixel contrast mismatch caused by manufacturing variation, with the $0.03$ standard deviation reflecting the ~$3\%$ variation reported for real sensors\cite{4444573}. The nominal value $C_{\text{nominal}}=0.3$, which was used during testing, matches the V2E default and sits near the midpoint of typical sensor thresholds\cite{hu_v2e_2021, delbruck2021feedbackcontroleventcameras}; it is exposed as a parameter so the pipeline can be tuned to a target sensor.

\subsection{Sensor Noise}
Two noise sources are modeled. Shot noise, arising from the quantal nature of photon arrival, is generated as ON/OFF events via a Poisson process consistent with V2E. Leak events are spontaneous ON events caused by junction leakage and parasitic photocurrent in the change-detector reset circuitry. Noise is applied after event generation and overrides any event previously written at the affected pixel. The full V2E noise model was not reproduced: the present implementation is a proof of concept aimed at validating the pipeline rather than maximizing fidelity, and additional effects such as the refractory period and hot pixels are left to future work.

\subsection{Automated Annotation}
The simulation components are packaged as a custom UE plugin, EventCamSimulation, which encapsulates the event-camera effect, its supporting materials and buffers, and three C++ modules implementing contrast-threshold sampling, leak events, and shot noise. A single manager actor placed in the scene drives the pipeline autonomously at runtime. EasySynth serves as the rendering backend and is the key to automated labeling: for each level sequence it renders, frame-by-frame, a paired DVS-style event frame (via the custom post-processing slot), a semantic-segmentation image, and an RGB frame, all written with identical filenames.

Annotation is then fully automated. A script pairs each event frame with its segmentation image by filename and remaps the event channels to the eTraM convention for compatibility with the eTraM detector. A second script extracts labels from the segmentation images using OpenCV: pixels are mapped to classes via the known class–color map, connected components per class are identified, and their axis-aligned bounding boxes are computed and converted to the normalized YOLO format, producing one label file per frame. The result is a fully paired event dataset with bounding-box annotations, generated without manual labeling. This is our central contribution: any UE level sequence can be converted into an annotated event dataset.

\subsection{Experimental Setup}
To illustrate the pipeline, $19$ level sequences were constructed ($7$ pedestrian-only, $5$ vehicle-only, $7$ mixed), each targeting a specific detection challenge: object scale (pedestrian traversals at distances of $30$, $15$, and $5$ m), motion speed, frontal versus rear appearance, partial occlusion, closely spaced and crossing instances, multi-class scenes with overlapping pedestrians and vehicles, ego-motion from a vehicle-mounted camera, and static elevated viewpoints. Night-time variants were produced by adjusting UE lighting (sun time, reduced sky/directional intensity, moonlight-blue tint) to test robustness across illumination changes. Together these form the UnrealEvent dataset, comprising $50{,}692$ annotated event frames ($34{,}013$ day, $16{,}679$ night). The roughly $2:1$ day-to-night ratio is deliberate, exposing the detector to varied lighting conditions that are especially relevant for event sensors operating on a logarithmic intensity scale.

The labeled synthetic data trains a YOLOv8 detector using the architecture established by eTraM for event-based detection, for $20$ epochs at batch size $64$. Generalization is assessed by evaluating the synthetically trained model on the real-world eTraM test set. All experiments are run with three random seeds and reported as mean$\pm$std; the primary metric is per-class AP50, matching the convention of the eTraM baseline, and all real-data evaluations use the validation-selected checkpoint.
\section{Results}
We evaluate the pipeline in three regimes, all summarized in Table~\ref{tab:results}: the model trained and tested on synthetic data; the same model applied zero-shot to the real eTraM test set; and the model after fine-tuning on a small real subset. The synthetic row is measured on the synthetic test set; all others are measured on the full static test split of eTraM ($30$ sequences; $110{,}224$ frames at a $33$\,ms accumulation window). Detection performance is reported as per-class AP50 (average precision at IoU $0.5$), the same metric as eTraM's published baseline.
\begin{table}[h!]
\centering
\caption{YOLOv8 detection performance across data regimes, reported as per-class AP50 (average precision at IoU $0.5$), mean$\pm$std over three seeds. $^{\dagger}$Values from the three-class (Pedestrian, Vehicle, Micro-mobility) real-data YOLOv8 baseline published by eTraM~\cite{verma_etram_2024}, its Micro-mobility class omitted. }
\label{tab:results}
\renewcommand{\arraystretch}{1.3}
\begin{tabular}{p{3.5cm} p{2.6cm} p{2.6cm}}
\toprule
\multicolumn{3}{c}{\textbf{YOLOv8 AP50 Results}} \\
\midrule
\textbf{Data Regime} & \textbf{Pedestrian} & \textbf{Vehicle}\\
\midrule
eTraM$^{\dagger}$ & $0.134$ & $0.314$\\
Synthetic data & $0.888{\pm}0.009$ & $0.991{\pm}0.002$\\
Zero-shot on eTraM & $0.059{\pm}0.019$ & $0.074{\pm}0.010$\\
Fine-tuned on eTraM & $0.356{\pm}0.012$ & $0.606{\pm}0.017$\\
\bottomrule
\end{tabular}
\end{table}

\subsection{Performance on Synthetic Data}
On the synthetic test set, which was $15\%$ of the generated data, the model reaches an AP50 of $0.991{\pm}0.002$ for vehicles and $0.888{\pm}0.009$ for pedestrians. The residual errors are missed detections rather than misclassifications: at the validation operating point, $97\%$ of vehicles but only $79\%$ of pedestrians are recovered, with the remainder assigned to background. This pattern is consistent with the sparse, edge-and-motion nature of event data, which provides less discriminative signal than dense frames, especially for small or partially occluded objects. The detections the model does make are reliable, qualitative examples carry confidence scores of $0.84-0.96$ with accurate localization, confirming that the limitation is recall, not classification or localization.\\
Vehicles outperform pedestrians across every metric, for two reinforcing reasons. First, vehicles occupy larger image regions and thus generate denser, more spatially coherent event patterns than pedestrians, who may span only a few pixels. Second, and more subtly, the training set carries a $2.55:1$ vehicle-to-pedestrian instance imbalance despite more pedestrian sequences being authored. This is a consequence of the sensing modality itself: because event cameras respond only to luminance change, slow-moving pedestrians at walking speed produce too few contrast changes to register consistently across accumulated frames, yielding far fewer valid annotated instances per sequence than vehicles. The imbalance is therefore an emergent property of event-based data, not a labeling artifact.

\subsection{Zero-Shot Generalization to eTraM}
If the synthetically trained model is directly applied to real eTraM data, performance drops sharply, to an AP50 of $0.074{\pm}0.010$ for vehicles and $0.059{\pm}0.019$ for pedestrians, with the model missing roughly $90 \%$ of pedestrians and $87 \%$ of vehicles. This result exposes a substantial sim-to-real domain gap, a well-documented obstacle in event-based vision\cite{articleJoubert, tan2025realcarlasdynamicvision}: synthetic event streams differ from real recordings in event density, noise statistics, and scene complexity.

\begin{figure}[h!]
\centering
\begin{subfigure}{0.33\textwidth}
  \includegraphics[scale=0.295]{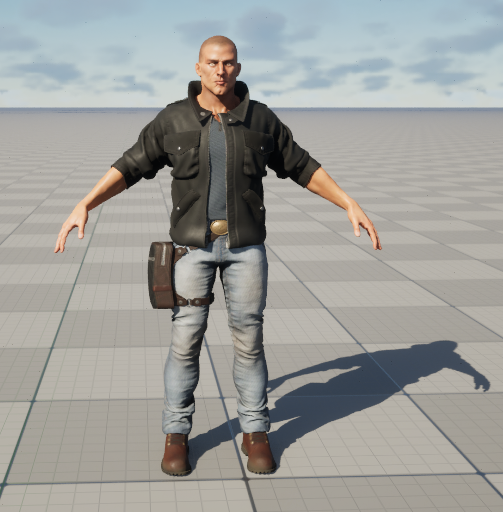}
    \caption{}
    \label{fig:PersonModelFront}
\end{subfigure}
\begin{subfigure}{0.15\textwidth}
  \includegraphics[scale=0.29]{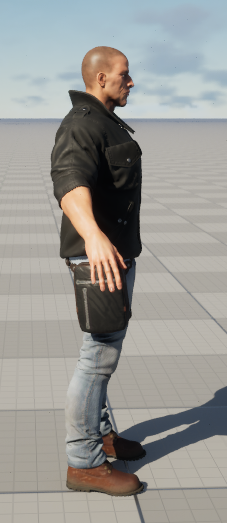}
  \caption{}
  \label{fig:PersonModelSide}
\end{subfigure}
\begin{subfigure}{0.17\textwidth}
  \includegraphics[scale=0.295]{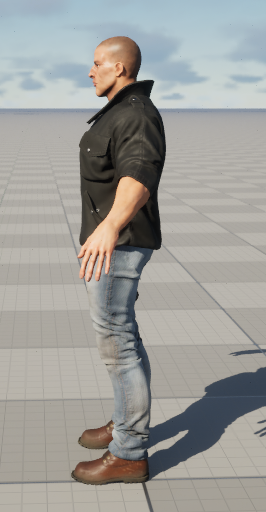}
  \caption{}
  \label{fig:PersonModelSide2}
\end{subfigure}
\begin{subfigure}{0.3\textwidth}
  \includegraphics[scale=0.32]{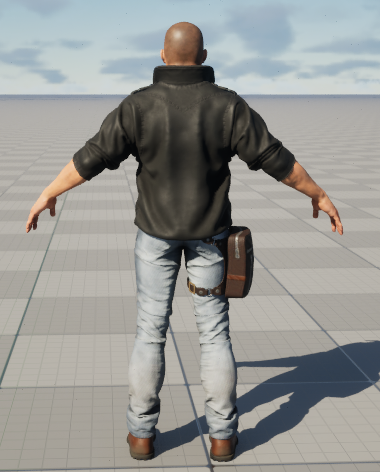}
  \caption{}
  \label{fig:PersonModelBack}
\end{subfigure}

\caption{Default pose of the pedestrian model in UE.}
\label{fig:PersonModel}
\end{figure}

Zero-shot transfer fails almost completely for both classes. After adapting the detector to the target domain, however, a concrete limitation of the synthetic data becomes apparent (Sec.~5.3): vehicles, which were rendered from six distinct models, recover much more of their performance after fine-tuning than pedestrians. In contrast, all pedestrian instances were generated from a single skeletal mesh with a default appearance, as shown in Fig.~\ref{fig:PersonModel}. As a result, the model was never exposed to the appearance and motion variability characteristic of real pedestrians. This suggests that limited asset diversity is an important constraint on pedestrian generalization.

Importantly, the qualitative results are more promising than the aggregate metrics alone would suggest. When the zero-shot model produces detections, they are typically assigned high confidence scores in the range of $0.55$ to $0.93$ for vehicles and $0.63$--$0.70$ for the sparse pedestrian detections, and are accompanied by spatially well-aligned bounding boxes. This indicates that the pipeline learns localization cues that transfer to some extent across domains, but remains poorly calibrated to the noise profile and accumulation characteristics of the target sensor.

\subsection{Fine-Tuning on a Small Real Subset}
\begin{figure}[ht]
    \centering
    \begin{subfigure}[b]{0.49\textwidth}
        \includegraphics[scale=0.2]{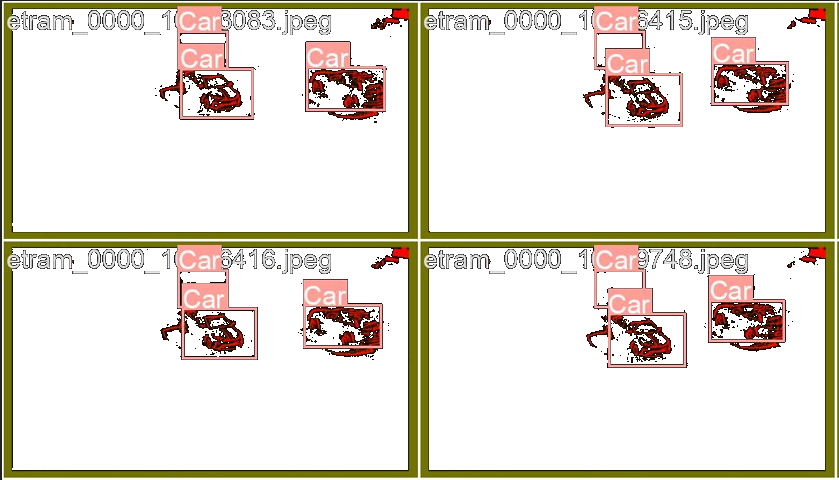}
    \end{subfigure}
    \begin{subfigure}[b]{0.49\textwidth}
        \includegraphics[scale=0.2]{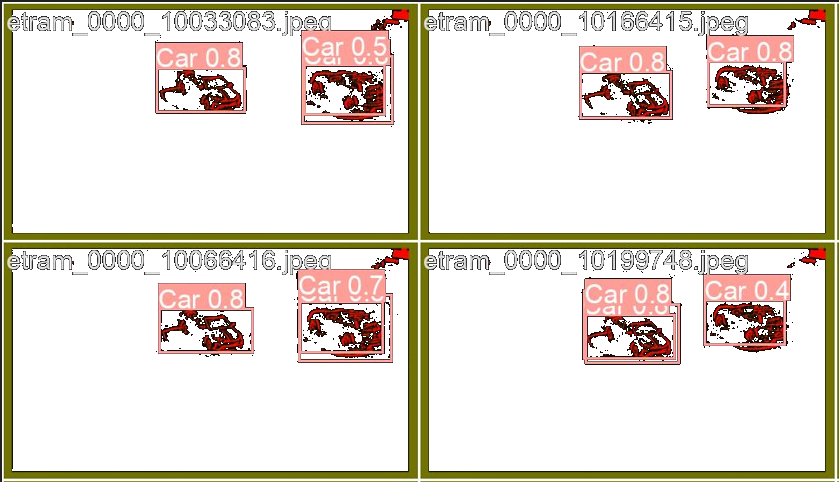}
    \end{subfigure}
    \begin{subfigure}[b]{0.49\textwidth}
        \includegraphics[scale=0.2]{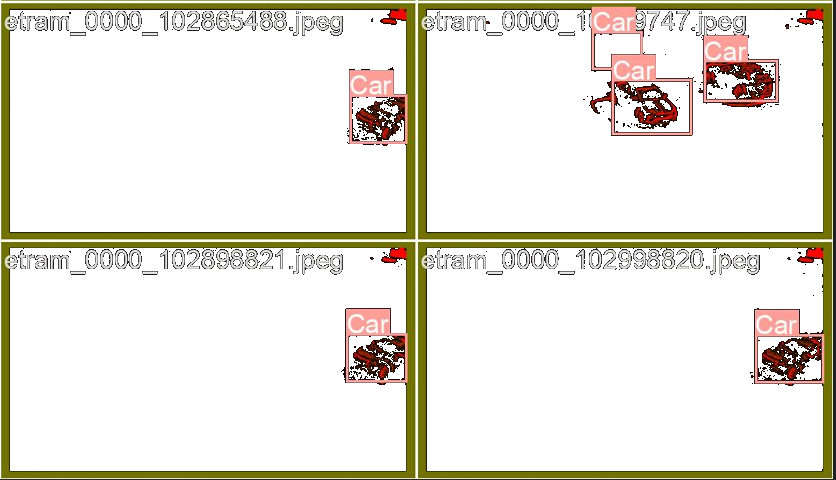}
    \end{subfigure}
    \begin{subfigure}[b]{0.49\textwidth}
        \includegraphics[scale=0.2]{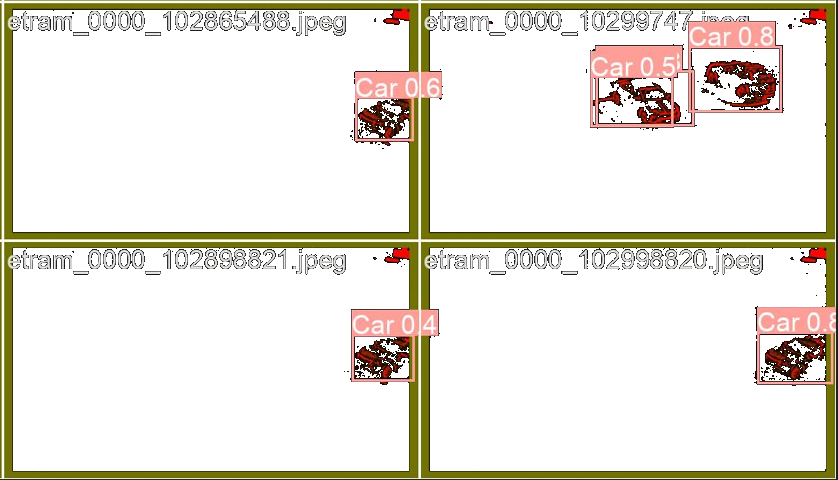}
    \end{subfigure}
\caption{Predicted eTraM test data using fine-tuned model.}
\label{fig:EtramTestData_finetuned}
\end{figure}

To adapt the detector to the target domain, we fine-tuned the synthetic-trained model on $2{,}436$ randomly selected eTraM training images spanning day and night, freezing the backbone so that synthetic-learned features are retained while the detection head adapts to the real distribution. Fine-tuning runs for a fixed $300$-epoch budget and we report the validation-selected checkpoint.
This improves real-domain performance: vehicle AP50 rises from $0.074$ to $0.606{\pm}0.017$ and pedestrian AP50 from $0.059$ to $0.356{\pm}0.012$, a 6-8x improvement over zero-shot, with the residual class gap consistent with the single-mesh limitation identified above. Qualitatively, predicted boxes align closely with ground truth and enclose vehicles with minimal offset; the main remaining defect is occasional duplicate boxes, again a recall-side rather than localization issue.
These results show that a small amount of real fine-tuning data substantially narrows the sim-to-real gap. However, the controlled comparison in Table~\ref{tab:ablation} indicates that this recovery is driven chiefly by the real data itself: with the real-data budget held fixed, initializing from the synthetic model performs on par with (full fine-tune) or slightly below (frozen backbone) training from scratch. The clearest paths to further improvement are more diverse pedestrian assets, a larger fine-tuning subset, and event-specific domain-adaptation methods. Representative results for the fine-tuned model are shown in Fig.~\ref{fig:EtramTestData_finetuned}.

\begin{table}[h!]
\centering
\caption{Controlled comparison on the real eTraM test set (mean$\pm$std over three seeds). AP50 here denotes the two-class mean; $\Delta$ is the per-seed paired difference in this quantity against the real-only baseline, with a $95\%$ bootstrap CI.}
\label{tab:ablation}
\small
\setlength{\tabcolsep}{3pt}
\renewcommand{\arraystretch}{1.2}
\begin{tabular}{l c c c}
\toprule
\textbf{Configuration} & \textbf{AP50} & \textbf{mAP@$\textbf{[.5{:}.95]}$} & \textbf{$\Delta$ AP50 [95\% CI]} \\
\midrule
\multicolumn{4}{l}{\emph{Baseline}} \\
\quad Real-only (from scratch) & $0.501{\pm}0.009$ & $0.216{\pm}0.004$ & --- \\
\midrule
\multicolumn{4}{l}{\emph{Synthetic pre-training (ours)}} \\
\quad Frozen backbone & $0.481{\pm}0.014$ & $0.204{\pm}0.005$ & $-0.020$ [$-0.031,-0.001$] \\
\quad Full fine-tune & $0.496{\pm}0.024$ & $0.217{\pm}0.010$ & $-0.005$ [$-0.040,+0.024$] \\
\bottomrule
\end{tabular}
\end{table}
%

\section{Future Work}
The pipeline was deliberately scoped as a proof of concept, using a stock UE 5.5 plugin to exercise the full generation, annotation, and evaluation workflow. This design made it possible to measure the remaining sim-to-real factors before addressing them through additional engineering. The controlled evaluation above suggests a clear ordering of future extensions by expected impact. Given that the fine-tuned pedestrian gap can be traced largely to the use of a single skeletal mesh, the most consequential improvement is greater asset diversity, including multiple pedestrian and vehicle models with varied appearance and motion. This directly targets the dominant source of sim-to-real error identified in our experiments. On the sensor side, the noise model can be extended beyond shot noise and leak events to include hot pixels and the refractory period\cite{hu_v2e_2021, nozaki_temperature_2017, articleJoubert}, both of which narrow the gap to physical sensors. Three further changes would broaden usability: making the output resolution configurable through a single parameter, so synthetic data can match a target sensor such as the Prophesee EVK4-HD ($1280\times720$) used by eTraM; exporting events in standard neuromorphic formats (AER, HDF5) rather than only image-based frames, preserving full temporal resolution for compatibility with established benchmarks; and making the temporal accumulation window adjustable, enabling denser representations aligned with the accumulation parameters of target datasets. Together these changes would let the pipeline produce data more closely matched to specific real-world sensors and processing frameworks.

\section{Conclusion}
This work presented UnrealEvent, a synthetic dataset generation and annotation pipeline for event-based vision built in UE 5.5. Addressing the research questions posed in the beginning: the pipeline generates high-quality annotated event data by coupling a custom V2E-style event effect with EasySynth's semantic rendering, from which bounding boxes are extracted automatically, so that any UE level sequence can be turned into a labeled event dataset without manual annotation. On generalization, a model trained predominantly on synthetic data and fine-tuned on a small real subset reaches an AP50 of $0.606$ for vehicles and $0.356$ for pedestrians on real eTraM data.


Rather than leaving the residual sim-to-real gap as a black box, the pipeline makes it measurable: because generation, annotation, and evaluation form a single controlled loop, individual factors can be isolated. The controlled comparison shows that, at the current level of event realism, synthetic pre-training does not yet yield a measurable benefit over training from scratch on the same real-data budget. This limitation appears to be caused by the specific assets and sensor parameters supplied to the pipeline, rather than from the generation-and-annotation approach itself. We then provides an evidence-ranked roadmap for improvement.

The principal practical cost is the UE expertise required to author diverse scenes; in return, the approach incurs no acquisition hardware cost, enables scenarios that are unsafe or difficult to capture physically, and scales through fully automated annotation. UnrealEvent therefore provides both a practical platform for controlled synthetic event-data generation and the measurement loop needed to close the gap it quantifies.
\section*{Acknowledgement}
This research was funded by the Flemish Government (AI Research Program).

\newpage
\bibliographystyle{splncs04}
\bibliography{main}
\end{document}